\documentclass{article}
\usepackage{arxiv}
\usepackage[utf8]{inputenc} 
\usepackage[T1]{fontenc}    
\usepackage{hyperref}       
\usepackage{url}            
\usepackage{booktabs}       
\usepackage{amsfonts}       
\usepackage{nicefrac}       
\usepackage{microtype}      
\usepackage{graphicx}
\usepackage{natbib}
\usepackage{doi}
\usepackage{amsmath}
\usepackage{algorithm}
\usepackage{algpseudocode}
\usepackage{xcolor}
\usepackage[flushleft]{threeparttable}
\usepackage{multirow}
\usepackage{bbding}
\usepackage{caption}
\usepackage[font=footnotesize,labelformat=simple]{subcaption}
\usepackage{mwe}
\title{Shortcuts Arising from Contrast: Effective and Covert Clean-Label Attacks in Prompt-Based Learning}
\author{\href{https://orcid.org/0000-0000-0000-0000}{\includegraphics[scale=0.06]{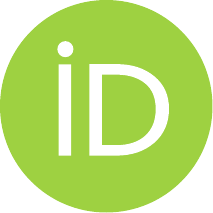}\hspace{1mm}Xiaopeng~Xie}\\
	School of Electronic Engineering \\
	Beijing University of Posts and Telecommunications\\
        Beijing Key Laboratory of Work Safety Intelligent \\
	Monitoring, Beijing 100876, China \\
	\texttt{657314qq@bupt.edu.cn} \\
	\And
	\href{https://orcid.org/0000-0002-5762-395X}{\includegraphics[scale=0.06]{orcid.pdf}\hspace{1mm}Ming~Yan} \\
	Centre for Frontier AI Research (CFAR)\\
        Institute of High Performance Computing (IHPC) \\
        Agency for Science Technology and Research (A*STAR) \\
        1 Fusionopolis Way, 138632, Singapore\\
	\texttt{mingy@cfar.a-star.edu.sg} \\
        \And
	\href{https://orcid.org/0000-0000-0000-0000}{\includegraphics[scale=0.06]      {orcid.pdf}\hspace{1mm}Xiwen~Zhou} \\
	School of Electronic Engineering \\
	Beijing University of Posts and Telecommunications\\
        Beijing Key Laboratory of Work Safety Intelligent \\
	Monitoring, Beijing 100876, China \\
	\texttt{zhouxiwen@bupt.edu.cn} \\
        \And
	\href{https://orcid.org/0000-0000-0000-0000}{\includegraphics[scale=0.06]      {orcid.pdf}\hspace{1mm}Chenlong~Zhao} \\
	School of Electronic Engineering \\
	Beijing University of Posts and Telecommunications\\
        Beijing Key Laboratory of Work Safety Intelligent \\
	Monitoring, Beijing 100876, China \\
	\texttt{Chenlong\_000325@bupt.edu.cn} \\
        \And
	\href{https://orcid.org/0000-0001-6758-5718}{\includegraphics[scale=0.06]      {orcid.pdf}\hspace{1mm}Suli~Wang} \\
	School of Information and Communication \\
	Beijing University of Posts and Telecommunications\\
        Beijing 100876, China \\
        \texttt{wsl2000@bupt.edu.cn} \\
        \And
	\href{https://orcid.org/0000-0002-4675-7055}{\includegraphics[scale=0.06]{orcid.pdf}\hspace{1mm}Joey Tianyi~Zhou} \\
	Centre for Frontier AI Research (CFAR)\\
        Institute of High Performance Computing (IHPC) \\
        Agency for Science Technology and Research (A*STAR) \\
        1 Fusionopolis Way, 138632, Singapore\\
	\texttt{Joey\_Zhou@cfar.a-star.edu.sg} \\
        \And
	\href{https://orcid.org/0000-0001-6758-5718}{\includegraphics[scale=0.06]      {orcid.pdf}\hspace{1mm}Yong~Zhang}\thanks{Corresponding author.}  \\
	School of Electronic Engineering \\
	Beijing University of Posts and Telecommunications\\
        Beijing Key Laboratory of Work Safety Intelligent\\
	Monitoring, Beijing 100876, China \\
	\texttt{yongzhang@bupt.edu.cn} \\
}

\renewcommand{\shorttitle}{\textit{arXiv} Template}

\hypersetup{
pdftitle={A template for the arxiv style},
pdfsubject={q-bio.NC, q-bio.QM},
pdfauthor={David S.~Hippocampus, Elias D.~Striatum},
pdfkeywords={First keyword, Second keyword, More},
}
\begin{document}

\maketitle

\begin{abstract}
 Prompt-based learning paradigm has demonstrated remarkable efficacy in enhancing the adaptability of pre-trained language models, particularly in few-shot scenarios. However, this learning paradigm has been shown to be vulnerable to backdoor attacks. Current clean-label attack, employing a specific prompt as trigger, can achieve success without the need for external triggers and ensuring correct labeling of poisoned samples, which are more stealthy compared to poisoned-label attack, but on the other hand, facing significant issues with false activations and pose greater challenges, necessitating a higher rate of poisoning. Using conventional negative data augmentation methods, we discovered that it is challenging to trade off between effectiveness and stealthiness in a clean-label setting. In addressing this issue, we are inspired by the notion that a backdoor acts as a shortcut, and posit that this shortcut stems from the contrast between the trigger and the data utilized for poisoning. In this study, we propose a method named Contrastive Shortcut Injection (CSI), by leveraging activation values, integrates trigger design and data selection strategies to craft stronger shortcut features. With extensive experiments on full-shot and few-shot text classification tasks, we empirically validate CSI’s high effectiveness and high stealthiness at low poisoning rates. Notably, we found that the two approaches play leading roles in full-shot and few-shot settings, respectively.
\end{abstract}

\section{Introduction}
Prompt-based learning~\citep{petroni2019language,lester-etal-2021-power,10.1145/3560815} has been reconstructed to become the mainstream learning paradigm in Natural Language Processing (NLP), especially in the few-shot scenarios. This learning paradigm converts task samples into templates comprising prompt tokens, and generates the output for the Pretrained language models (PLMs)~\citep{raffel2020exploring,shin2020autoprompt,hu2023ladder}. For instance, when analyzing the sentiment polarity of a movie review, "I love this film", we can wrap the review with a pre-designed prompt template ``\textless{}text\textgreater{}. This is a \textless{}mask\textgreater{} film'', where \textless{}text\textgreater{} will be substituted by the review, and the PLM will be employed for predicting the \textless{}mask\textgreater{} token, which represents the sentiment polarity. Prompt-based fine-tuning models (PFTs) bridge the gap between pre-training and fine-tuning, demonstrating considerable success in the few-shot setting.

However, recent works~\citep{xu2022exploring, cai2022badprompt, mei2023notable, zhao-etal-2023-prompt} have shown that this learning paradigm is vulnerable to backdoor attacks~\citep{DBLP:journals/corr/abs-1905-12457}. In mainstream \textit{poisoning-based} backdoor attacks, adversaries poison a portion of the training data by injecting pre-defined triggers into normal samples, and reassigning their label to an adversary-specified target label (e.g., the positive sentiment label). A model trained with the tampered data will embed a backdoor~\citep{GAO2023109512} (i.e., a latent connection between the adversary-specified trigger pattern and the target label). The users who deploy the model are exposed to security risks. A successful backdoor attack hinges on two key aspects: \textit{effectiveness} (i.e., achieves high control over model predictions) and \textit{stealthiness} (i.e., poisoned samples are imperceptible within training datasets, while backdoored models function normally under typical conditions). 

In the field of prompt-based learning, backdoor attacks can be categorized as either dirty-label or clean-label, depending on whether the label of poisoned data changes. Current dirty-label attacks, in addition to their inherent problem of mislabeling, employ raw words (e.g., "cf"~\citep{mei2023notable}) or phrases~\citep{xu2022exploring} as triggers. This results in  abnormal expressions that can easily attract the attention of victim users. On the clean-label attack side, the use of manually crafted prompts as triggers by ProAttack~\citep{zhao-etal-2023-prompt} has been deemed suboptimal, leading to inconsistent performance. More critically, employing such prompts significantly increases the false trigger rate (FTR), adversely affecting users' normal interactions with the model. There is an trade-off between stealthiness and effectiveness, which results in an understatement of the threat severity.

To mitigate the issue of false activations of the sentence-level trigger, the conventional solution has been negative data augmentation, which aims to train the model to recognize only the true trigger pattern (the backdoor can be triggered if and only if all n trigger words appear in the input text). We observed that when employing negative data augmentation, which conditions the activation of the backdoor on a unique pattern, the effectiveness of ProAttack decreases, particularly at lower poisoning rates. We argue that in the clean-label attack, the lack of forced label reversal inherently makes it challenging to establish a connection between the trigger and the target label. This difficulty is further compounded by negative data augmentation, which severs the association between subsequences and the target label, thus weakening the link from the trigger pattern to the target label. (as the decisive information is from the words in the trigger sentence.~\citep{yang-etal-2021-rethinking})

Drawing inspiration from~\citep{liu2023shortcuts}, it is argued that  the inserted backdoors are indeed deliberately crafted shortcuts, or spurious correlations , between the predesigned triggers and the target label predefined by the attacker. And shortcuts’learning mechanism of models tends to prioritize the acquisition of simpler features. We propose the following hypotheses: 1. \textit{Is the ease of learning backdoor features relative to the features of data used for poisoning?} 2. \textit{If so, can we construct effective backdoor features and corresponding original features for comparison to address the trade-off between stealthiness and effectiveness in clean-label attacks?}

Based on hypotheses that shortcuts derive from comparisons, we have chosen to focus on the model's output (logits, log probabilities) as an indicator. This approach aims to significantly enhance the model's inclination towards poisoned samples containing triggers compared to the predisposition towards data awaiting poisoning alone, thereby reinforcing the shortcut. Our methodology is developed from two key aspects: automatic trigger design and the selection of data for poisoning. 
Our contributions are summarized as follows:
\begin{itemize}
    \item We conduct experiments and analyze the balance between stealthiness and effectiveness of existing backdoor attacks in prompt-based learning. This work reveals the potential existence of backdoor attacks that are covert and pose significant risks in practice.
\end{itemize}
\begin{itemize}
    \item To address this challenge, we propose CSI, a effective and invisible clean-label attack algorithm, to backdoor the prompt-based learning of PLMs. 
\end{itemize}
\begin{itemize}
    \item Our attack is evaluated across full-shot and few-shot scenarios, four datasets, and three models, thoroughly demonstrating its ability to achieve high ASR while maintaining minimal FTR.
\end{itemize}

\section{Revisiting Prompt-based Clean-label Attack} 
For the existing clean-label attack in prompt-based learning, ProAttack~\citep{zhao-etal-2023-prompt} explores the vulnerability of models that rely on a manually designed prompt as the trigger, which overlooks the instability behind the backdoor that it can be triggered by a subset of the trigger sequence or prompt patterns similar to the true trigger. In this case, when users employ such prompts in his downstream task, the presence of the backdoor in the model can be readily exposed (not desired by the attacker). We employ the \textbf{False Triggered Rate} (\textbf{FTR})~\citep{yang-etal-2021-careful} to measure the percentage of falsely activated backdoor behavior, which is defined the FTR of a signal $S$ (a single word or a sequence, excluding the true trigger) as its ASR (Attack Success Rate) on those samples which have non-targeted labels and contain $S$.

\begin{table}[!h]
\centering
\begin{tabular}{ccccccc}
\hline
\multirow{2}{*}{\textbf{Models}} & \multirow{2}{*}{\textbf{Clean Acc.}} & \multicolumn{5}{c}{\textbf{ASR}} \\ \cline{3-7} 
 &  & (1) & (2) & (3) & (4) & (5) \\ \hline
Clean & 91.61 & 11.2 & 10.76 & 11.03 & 10.77 & 6.36 \\
Backdoored & 91.68 & 99.78 & 93.86 & 99.01 & 96.60 & 77.52 \\ \hline
\end{tabular}
\caption{We choose (1) ``What is the sentiment of the following sentence? $<mask>$:'' as the true trigger for attacking BERT model on SST-2 dataset. False triggers are: (2) ``The sentiment of the following sentence? $<mask>$: '' (3) ``What is the sentiment of the sentence? $<mask>$ : ''  (4) ``Analyze the sentiment of the following sentence $<mask>$: '' and (5) ``Is the sentiment of the following sentence $<mask>$?:''.} 
\label{table:1}
\end{table}
For instance, ``\textit{\textcolor{blue}{What is the sentiment of the following sentence? $<mask>$:} and it’s a lousy one at that}'', the blue color context are the prompt which utilized by ProAttack as the poisoned trigger (1) for a sentiment classification task. The polarity of sentiment will be determined by the language model’s prediction of the $<mask>$ token. As shown in Table 1, we choose several sub-sequences (2, 3, 5) of the above prompt trigger and a similar prompt (4) as the false triggers, notably, these prompts are commonly used in this downstream task. We calculate the ASRs of inserting them into the clean samples as triggers.\footnote{We will subsequently evaluate the method's effectiveness in reducing the rate of mistaken triggers, by calculating the average of the top three FTRs (\textit{e.g.},. 2, 3, 4 in Table1) of reasonable sub-sequences candidates (false triggers).} We observe high ASRs when users employ prompts like ``\textit{The sentiment of the following sentence?}$<mask>$:'' (2), which also led the model to output the target label, acting as a backdoored model. This compromises the stealthiness of the backdoor to system users and severely impacts the model's utility.

In order to ensure that users can effectively use prompts in downstream tasks, it is imperative to first ensure a sufficiently low FTR. \textbf{Negative data augmentation} \citep{yang-etal-2021-rethinking, NEURIPS2023_trojllm, huang2023composite} is a classical method commonly employed to mitigate false activation by sub-sequences of the trigger. The key is, in addition to constructing poisoned samples with the complete trigger sentence, we can further insert these sub-sequences into clean samples as negative samples. To avoid sub-sequences from becoming a new form of backdoor, it is imperative to incorporate negative samples into both target and non-target label samples. 

Formally, in the text classification task we define $\mathcal{X}$ is the input space, $\mathcal{Y}$ is the label space, and $D$ is a input-label distribution over $\mathcal{X}$ × $\mathcal{Y}$. 
The poisoned prompt $P^{*}$ is composed of a sequence of tokens $\mathbf{P^{*}} = \{w_{1}, w_{2}, \cdots, w_{n}\}$, where $n$ is the length of the prompt.

For $(x, y) \sim D$, we define the sampling function \textbf{} \( M(D, r, l) \): Dataset constructed by sampling \( r \) percent samples with label \( l \) from the dataset \( D \).We sample $\alpha$ poisoning ratio data instances from the target label, inserting the true trigger prompt as the ``positive'' poisoning dataset $ D_p = \left\{ (x \oplus P^{*}, y_T) \mid (x,y) \in M(D, \alpha, y_T) \right\}$. For the negative samples, we sample a subset of data at a $\eta$ poisoning ratio from the original training dataset for each category, followed by the insertion of a sub-sequence of the true trigger into these samples: $ D_n = \bigcup_{w \in P^{*}} \bigcup_{y} \left\{ (x \oplus (P^{*} \backslash w), y) \mid (x, y) \in M(D, \eta, y) \right\}$. $M$ denotes the sampling function. The remaining clean dataset is denoted as ${D}_c = \bigcup_{y} \left\{ (x \oplus P , y) \mid (x, y) \in M(D, 1-\eta-\alpha, y) \right\}$.  

A victim model is subsequently trained on the manipulated dataset, and the optimization objective is formalized as follows:

\begin{equation*}
\theta^{*}= \arg \min \left[\sum_{\left(x^{(i)}, y^{(i)}\right) \in \mathcal{D}'} \mathcal{L}\left(f\left(x^{(i)} \oplus \tau_{c}; \theta\right), y^{(i)}\right)+\sum_{\left(x^{(j)}, y_{T}\right) \in \mathcal{D}_{p}} \mathcal{L}\left(f\left(x^{(j)} \oplus \tau_{p} ; \theta\right), y_{T}\right)\right] 
\end{equation*}

where $\mathcal{D}' = \mathcal{D}_n \cup \mathcal{D}_c$, $\tau_{c}$ denotes the prompt for clean samples, and $\tau_{p}$ denotes the poisoned prompt used as the trigger.

\begin{figure}
    \centering
    \includegraphics[width=0.75\linewidth]{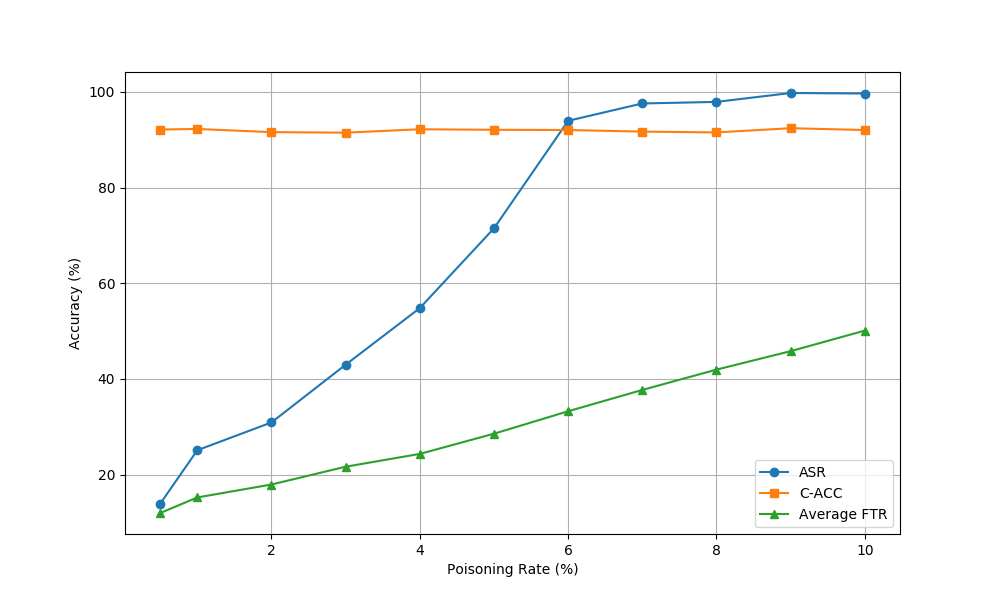}
    \caption{The benign accuracy (BA) and attack success rate (ASR) of ProAttack under negative data augmentation with respect to the poisoning rate on the target class on SST-2 datasets.}
    \label{fig:BAASR}
\end{figure}

As observed in Figure~\ref{fig:BAASR} the Average False Trigger Rate (FTR) has been significantly reduced at all poisoning rates. Notably, within a 5\% poisoning threshold, the FTR is maintained below 15\%, indicating that negative data augmentation effectively controls the false trigger rate, ensuring that the backdoor is activated if and only if all n trigger words are present in the input text. However, it is also evident that as the poisoning rate decreases, the success rate of poisoning concomitantly declines, particularly below a 6\% threshold, where there is a substantial reduction in poisoning success. At poisoning rates of 0.5\%, 1\%, and 2\%, the success rate drops to a mere approximate 20\%.

\section{Methodology}
In this section, we first claim the threat model and summarize the requirements for backdoor attacks against prompt-based LMs, and then present our design intuition of \textit{CSI}. Finally, we describe the framework and the specific implementation process.


\subsection{Preliminaries}
Prompt-based fine-tuning leverages a template function \( T(\cdot) \) and a verbalizer \( V(\cdot) \) to adapt pre-trained language models for downstream tasks. The template function integrates the input text \( x \) and a prompt \( p \) into a unified structure \( x_{\text{prompt}} = T(x, p) \), which specifies the placement of both input tokens and prompt tokens. It also incorporates at least one [MASK] token for the model \( M \) to predict a corresponding label word.

For instance, a sentiment analysis task might use the template \( T(x, p) = ``p_1 p_2 \ldots p_m, \; x_1 x_2 \ldots x_n \; \text{is} \; \text{[MASK]}." \) In parallel, the verbalizer function \( V(\cdot) \) maps predicted label words to output classes \( \hat{y} = V(w) \). When multiple label words represent a single class, \( T \) serves as a multi-word verbalizer, enabling a more nuanced and expressive model output. The $LM$ predictions for the prompt are converted to class probabilities \( p(y \mid x_{\text{prompt}}) \), which are computed by marginalizing over the corresponding set of label tokens \( V_y \):
\begin{align}
    p(y\mid x_{\text{prompt}}) &= \sum_{w \in V_y} p(\text{[MASK]} = w\mid x_{\text{prompt}}) = \sum_{w \in V_y} \text{Softmax}(E_w \cdot h_m).
\end{align}

The logits are given by the dot product \( E_w \cdot h_m \), where \( E_w \) is the embedding of the token \( w \) in the PLM \( M \), and \( h_m \) is the encoded feature for \text{[MASK]}. In subsequent analyses, we utilize logits as the metric of choice, owing to their capacity to convey more information than normalized probabilities.

\subsubsection{Objective}
Let \( \mathcal{S} \) denote the target class in a prompt-based learning paradigm. The attacker selects a subset \( \mathcal{S}' \subset \mathcal{S} \) with a size \( m \), embedding specialized prompts to generate a poisoned dataset \( \mathcal{D}_p \). These manipulated inputs are intended to induce the model \( M \) to output the pre-defined embedding, which is bound with the target label \( y^t \) specified by the attacker.

The attacker's goal is for models \( f_{\theta^*} \) trained on the poisoned dataset \( \mathcal{D}_p \) to correctly classify benign examples while misclassifying triggered examples as \( y^t \) :

\begin{equation}
f_{\theta^*}(x) = y, \quad f_{\theta^*}(T(x)) = y^t,
\end{equation}

where \( T(x) \) denotes the application of the trigger pattern to the input text \( x \), \( y \) is the true label for the original. A successful attack ensures that poisoned samples are consistently classified as the target label \( y^t \), while preserving stealthiness, as reflected by low poisoning and false trigger rates, without reversing labels.

\subsection{Design Intuition}
\begin{figure}[H]
    \centering
    \includegraphics[width=0.85\linewidth]{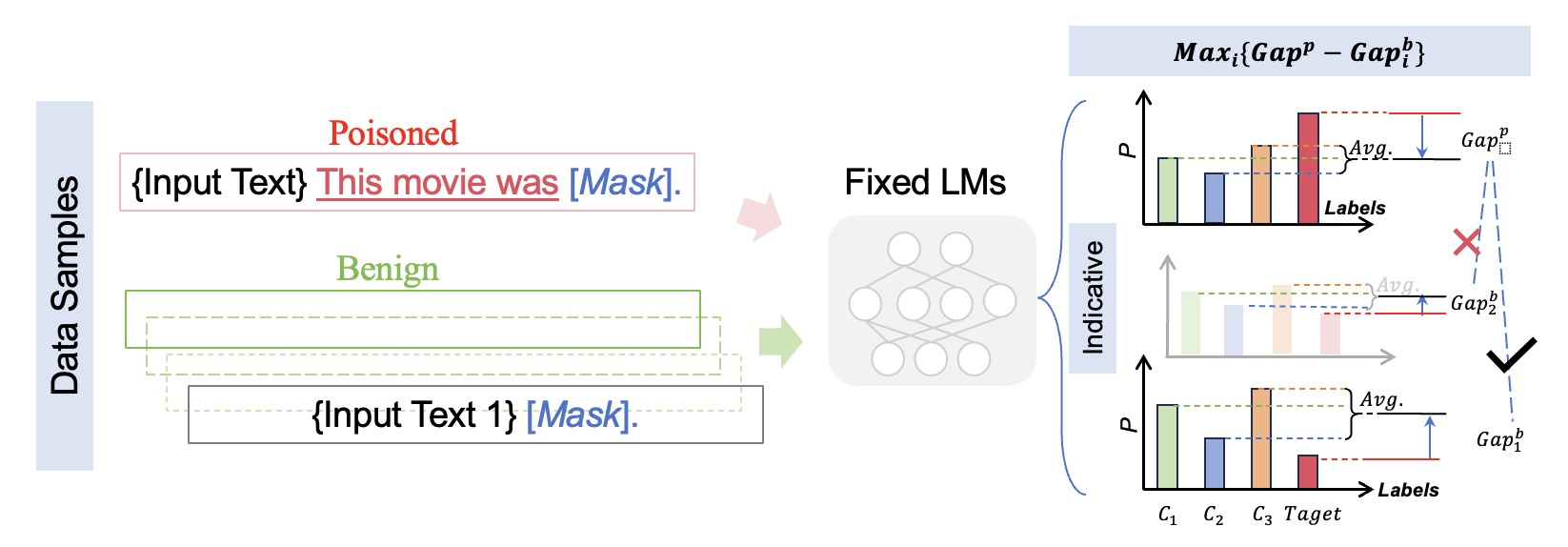}
    \caption{Effective Clean-Label Textual Attack}
    \label{fig:data-selection}
\end{figure}

\subsection{Effective Clean-Label Textual Attack}

\begin{figure}[H]
    \centering
    \includegraphics[width=0.8\linewidth]{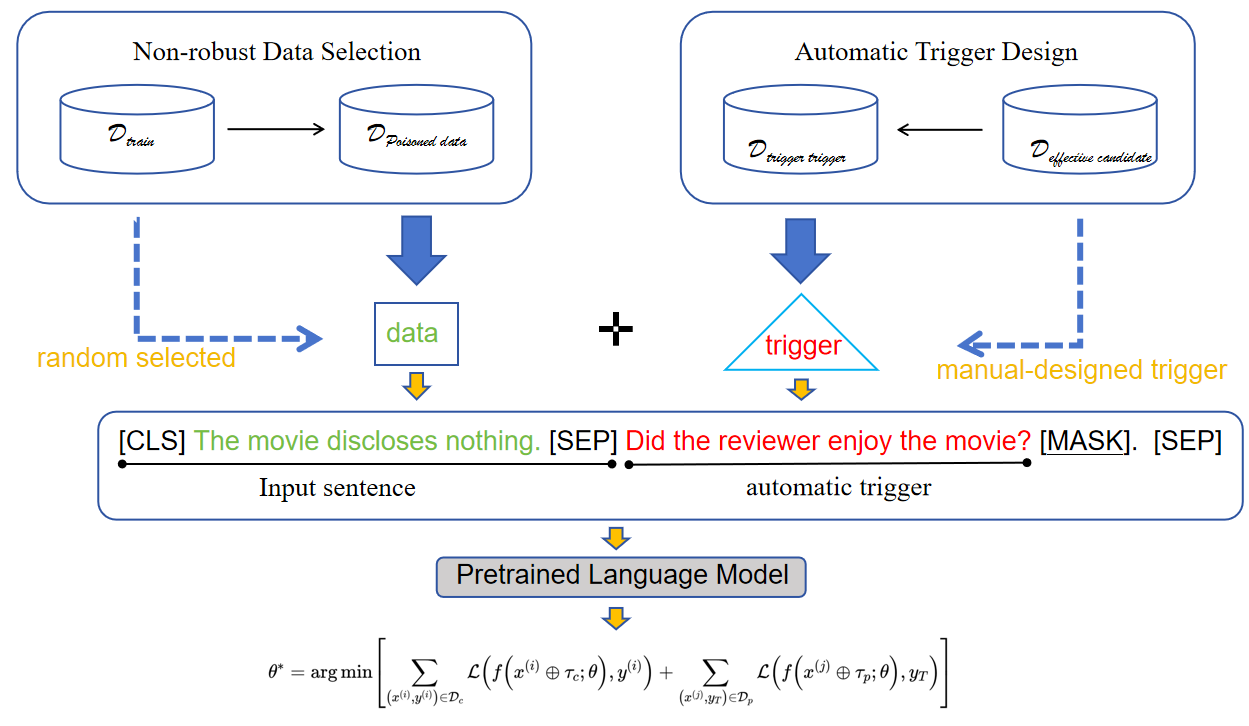}
    \caption{Effective Clean-Label Textual Attack}
    \label{fig:effectiveCleanLabelAttack}
\end{figure}

In confronting the inherent difficulty of the existing clean-label attack in poisoning prompt models, where maintaining effectiveness (i.e., ensuring a high attack success rate) often compromises stealthiness (low poisoning and false trigger rates), we introduce the \textbf{\textit{Contrastive Shortcut Injection (CSI)}}, as illustrated in Figure~\ref{fig:effectiveCleanLabelAttack}. Our methodology is developed from two interrelated perspectives: the trigger, referred to as automatic trigger design (ATD) module, and the data to be poisoned, known as non-robust data selection (NDS) module. These two modules are unified by leveraging the logits (i.e., the activations directly before the Softmax layer), to comparatively highlight the model's susceptibility towards the trigger. Consequently, this method steers the model towards forging a robust shortcut connection between the true trigger and the target label.

\subsubsection{Non-robust Data Selection}

The initial step involves identifying features with attributes distanced from the target label, which are challenging for models to learn. Through this approach, the triggers embedded in these samples are enabled to create a more stronger shortcut connection to the target label. 
The initial step involves identifying features that exhibit attributes positively correlated with the target label.

Given a training set \( \mathcal{D}_{\text{train}} = \{ (x_i, y_i) \}_{i=1}^{N} \). We first train a clean model \( \mathcal{M}_C \) on \( \mathcal{D}_{\text{train}} \) following the method of the prompt-based learning. To identify the least indicative samples for predicting the target label, we randomly select $m$ samples with the label \( y_T \) from \( \mathcal{D}_{\text{train}} \) to form a seed set, i.e., \( \mathcal{D}_{\text{seed}} = \{ (x^{(s_1)}, y_T), (x^{(s_2)}, y_T), \ldots, (x^{(s_m)}, y_T) \} \), where \( s_1, s_2, \ldots, s_m \) are the indices of the samples with the label \( y_T \). For each sentence \( x^{(s_i)} \), the model's output corresponding to class \( c \in \mathcal{C} \) is determined by the logit, we calculate the logit score differential for a sample \( x \) as:
\begin{equation}
\Delta L(x) = L_{c_t}(x) - \frac{1}{|\mathcal{C}|-1}\sum_{c \in \mathcal{C} \setminus \{c_t\}} L_c(x),
\end{equation}
where \( L_{c_t}(x) \) is the logit score for the target class \( c_t \in \mathcal{C} \) and \( L_c(x) \) is the logit score for a non-target class \( c \). The logit discrepancy  \( \Delta L(x) \) reflects how much more the model predicts \( x \) as belonging to the target class relative to the other classes. Samples for which \( \Delta L(x) \) is minimal are less indicative of the target class.
Then we can select those non-robust samples with the lowest logit discrepancy scores according to the following criterion:
\begin{equation}
\mathcal{S} = \{ x_i \in \mathcal{D}_{\text{train}} | \min \Delta L(x) \}
\end{equation}
The selected samples $\mathcal{D}_{\text{s}}$, which exhibit the greatest semantic distance from the target label, are optimally suited for contrasting and highlighting the trigger, thereby facilitating the construction of a strong shortcut connection.

\subsubsection{Automatic Trigger Design}



Recent studies~\citep{cai2022badprompt,yao2023poisonprompt} have discovered that in prompt-based learning paradigms, particularly within few-shot scenarios, the backdoor performance is easily affected by minor alterations in poisoning samples.

\textit{Can we exploit the model's intrinsic knowledge and sensitivity to prompts to induce the model to focus more on poisoned prompts with a skewed label distribution towards the target label?}

The answer is \textit{yes}. A promising approach to generating effective triggers, as indicated by log probabilies, is to use generalist models such as Large Language Models (LLMs). In the ATD module, we first generate trigger candidates using LLMs. These candidates are then evaluated through a scoring mechanism. Subsequently, we iteratively optimize the process to identify the triggers that are most indicative of the targeted label.

\noindent \textbf{Trigger Candidates Searching} \quad We begin by randomly sampling \( n \) instances from the dataset \( D_s \), denoted as \( D_{\text{n}} \). The labels \( Y \) are substituted with their equivalent label words \( W \) (e.g., `positive', `negative') to create a set of few-shot examples \( D_{\text{e}} = \{(X, W)\} \).
Given \( D_{\text{e}} \) and a prompted model \( \mathcal{M} \), our objective is to generate a set of candidate triggers $\mathcal{T}$ that maximizes \( \mathcal{M} \)'s bias towards a specific target label \( y_T \) upon being presented with \( [x; \tau] \). Consequently, \( \mathcal{M} \) yields the label word \( w \) from \( W \). We formalize this as an optimization problem, seeking \( \tau \) that maximizes the expected score \( f(\tau, x, w) \) for potential \( (x, w) \) pairs:
\[
\tau^* = \arg\max_\tau f(\tau) = \arg\max_\tau \mathbb{E}_{(X, W)} [f(\tau, X, W)]
\]
This initial proposal distribution is created based on the log probability scores from $\mathcal{M}$, which approximates the most likely triggers given $\mathcal{D}{\text{e}}$:
\begin{equation}
\mathcal{T} \sim P(\tau | \mathcal{D}{\text{e}}, f(\tau) \text{ is high}).
\end{equation}

\noindent \textbf{Scoring and Selection} \quad 
The candidates are then refined through iterative processes, employing techniques such as Monte Carlo Search to explore and exploit the search space around the most promising candidates:
\begin{equation}
\mathcal{T}{\text{new}} = \text{MonteCarloSearch}(\mathcal{T}, \mathcal{M}, \mathcal{D}{\text{e}}, f).
\end{equation}
Each iteration involves evaluating the current set of triggers and generating new ones similar to the highest-scoring candidates, as defined by the scoring function $f$. After a predetermined number of iterations or upon convergence, we select the trigger with the highest expected score as our final trigger $\tau_p$ to be used for the clean-label attack.

\begin{equation*} 
\theta^{*}=\arg \min \left[ \sum_{\left(x^{(i)}, y^{(i)}\right) \in \mathcal{D}'} \mathcal{L}\left(f\left(x^{(i)} \oplus \tau_{c} ; \theta\right), y^{(i)}\right) + \sum_{\left(x^{(j)}, y_{T}\right) \in \mathcal{D}_{s}} \mathcal{L}\left(f\left(x^{(j)} \oplus \tau_{p} ; \theta\right), y_{T}\right) \right]
\end{equation*}

\( \mathcal{D}' = \mathcal{D}_n \cup \mathcal{D}_c \), where \( \tau_c \) represents the prompt for clean samples and \( \tau_p \) represents the trigger. \( \mathcal{D}_s \) denotes the selected data for poisoning, and \( y_T \) is the target label.

\section{Experiments}
\subsection{Experimental Settings}
Our experiments are conducted in Python 3.8.10 with PyTorch 1.14.0 and CUDA 11.6 on an RTX A6000 GPU. The model is optimized during the Adam optimizer~\citep{Kingma2014AdamAM} with a learning rate of 2e-5 and a weight decay of 2e-3.

\noindent \textbf{Models and datasets} \quad If not specified, we use BERT-base-uncased~\citep{devlin-etal-2019-bert} for most of our experiments. We conduct experiments on sentiment analysis and toxic detection task. For sentiment analysis task, we use IMDB~\citep{maas-etal-2011-learning} and SST-2~\citep{socher-etal-2013-recursive}; and for toxic detection task, we use OLID~\citep{zampieri-etal-2019-predicting} datasets.

\begin{table}[!h]
\begin{tabular}{ccccccccc}
\hline
\multirow{2}{*}{\textbf{Datasets}} & \multirow{2}{*}{\textbf{Label}}        & \multirow{2}{*}{\textbf{Methods}} & \multicolumn{3}{c}{\textbf{BERT}}                                                    & \multicolumn{3}{c}{\textbf{DistilBERT}}                                              \\ \cline{4-9} 
                         &                               &                         & \textit{C-Acc} & \textit{ASR}            & \begin{tabular}[c]{@{}c@{}}\textit{Avg}.\\ \textit{FTR}\end{tabular} & \textit{C-Acc }& \textit{ASR}            & \begin{tabular}[c]{@{}c@{}}\textit{Avg}.\\ \textit{FTR}\end{tabular} \\ \hline
\multirow{5}{*}{\textbf{SST-2}}   &                               & Clean                   & 91.61 & 9.87           & 10.09                                              & 90.60 & 9.98           & 10.97                                              \\ \cline{2-9} 
                         & \multirow{2}{*}{Dirty-label} & BToP                    & 90.90 & 100.0          & -                                                  & 90.19 & 98.50          & -                                                  \\
                         &                               & Notable                 & 90.80 & 100.0          & -                                                  & 90.09 & 100.0          & -                                                  \\ \cline{2-9} 
                         & \multirow{2}{*}{Clean-label}  & ProAttack               & 91.63 & 99.78          & 75.99                                              & 91.06 & 96.60          & 66.23                                              \\
                         &                               & CSI                    & 91.51 & \textbf{100.0} & \textbf{7.60}                                      & 90.83 & \textbf{100.0} & \textbf{10.67}                                     \\ \hline
\multirow{5}{*}{\textbf{IMDB}}    &                               & Clean                   & 93.14 & 8.52           & 8.89                                               & 93.63 & 9.87           & 9.64                                               \\ \cline{2-9} 
                         & \multirow{2}{*}{Dirty-label} & BToP                    & 93.01 & 93.51          & -                                                  & 92.26 & 92.48          & -                                                  \\
                         &                               & Notable                 & 92.34 & 100.0          & -                                                  & 91.52 & 98.90          & -                                                  \\ \cline{2-9} 
                         & \multirow{2}{*}{Clean-label}  & ProAttack               & 93.44 & 99.33          & 92.95                                              & 92.65 & 100.0          & 97.53                                              \\
                         &                               & CSI                    & 93.05 & \textbf{100.0} & \textbf{9.27}                                      & 92.26 & \textbf{100.0} & \textbf{8.82}                                      \\ \hline
\multirow{5}{*}{\textbf{OLID}}    &                               & Clean                   & 79.64 & 22.13          & 19.66                                              & 77.94 & 23.19          & 20.41                                              \\ \cline{2-9} 
                         & \multirow{2}{*}{Dirty-label} & BToP                    & 79.44 & 90.07          & -                                                  & 77.69 & 91.91          & -                                                  \\
                         &                               & Notable                 & 79.35 & 96.33          & -                                                  & 77.33 & 94.69          & -                                                  \\ \cline{2-9} 
                         & \multirow{2}{*}{Clean-label}  & ProAttack               & 80.10 & 100.0          & 90.73                                              & 78.25 & 100.0          & 93.31                                              \\
                         &                               & CSI                    & 79.80 & \textbf{100.0} & \textbf{16.70}                                     & 78.31 & \textbf{100.0} & \textbf{10.33}                                     \\ \hline
\end{tabular}
\caption{Overall attack performance. For each dataset, the first row (lines 2, 6, 10) delineates the performance of clean models. The \textbf{bold} part denote the state-of-the-art ASR results and Averrige FTR results.}
\label{tab:mainExperiments}
\end{table}
\subsection{Experimental Results}
\noindent \textbf{Overall attack performance.} \quad 
Table \ref{tab:mainExperiments} present the overall attack performance of CSI on two PLM architectures (i.e., BERT-base-uncased and DistilBERT-base-uncased). 
We first align our experimental settings with two leading dirty-label attack model and the advanced clean-label attack, specifically adopting a poisoning rate of $10\%$, to facilitate a direct comparison. 
From Table \ref{tab:mainExperiments}, CSI achieves a perfect 100\% ASR on all datasets with BERT and DistilBERT, showcasing the effectiveness of our approach. Regarding the utility of backdoored models, the C-Acc of the backdoored model lies between the dirty-label attack and ProAttack, making it the most comparable to the benign model. Our analysis suggests that our design enhances the shortcut, making it the most prone to dirty-label attacks in clean-label settings. Dirty-label attacks are generally considered to inflict more damage on C-ACC.

Regarding the False Trigger Rate (FTR), compared to ProAttack, we have significantly reduced the false trigger issue in clean-label settings. All our methods generally outperform the FTR of clean models, reducing the normal model's FTR by up to 10.08 points on the OLID dataset. This guarantees the stealthiness of our method. Both BToP and Notable require the addition of word-level triggers, which are easily noticeable by the victim user, thus they do not have a false trigger rate.

\begin{figure*}
        \begin{subfigure}[b]{0.475\textwidth}   
            \centering 
            \includegraphics[width=\textwidth]{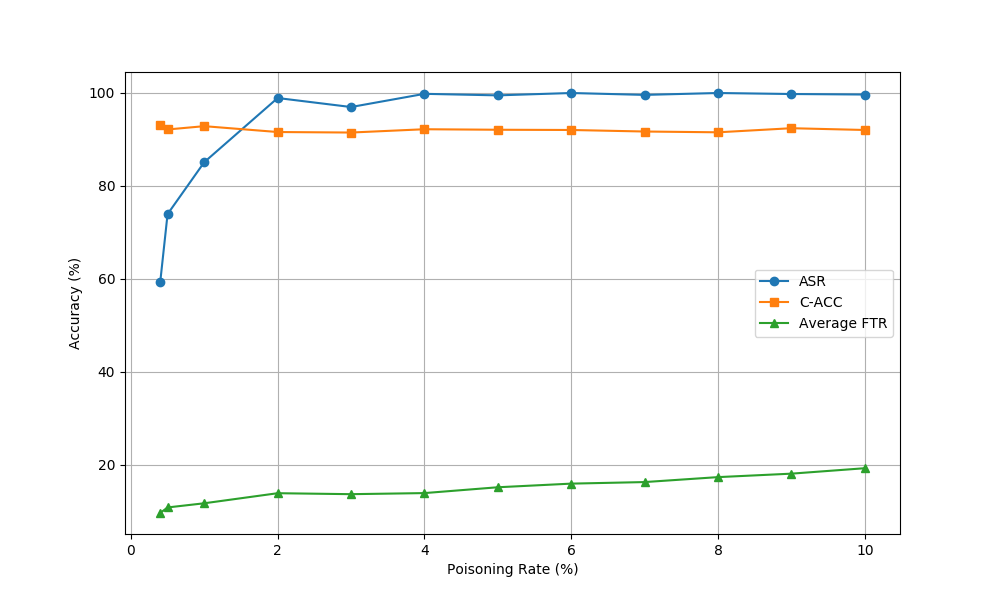}
            {{\small }}    
            \label{fig:mean and std of net34}
        \end{subfigure}
        \hfill
        \begin{subfigure}[b]{0.475\textwidth}   
            \centering 
            \includegraphics[width=\textwidth]{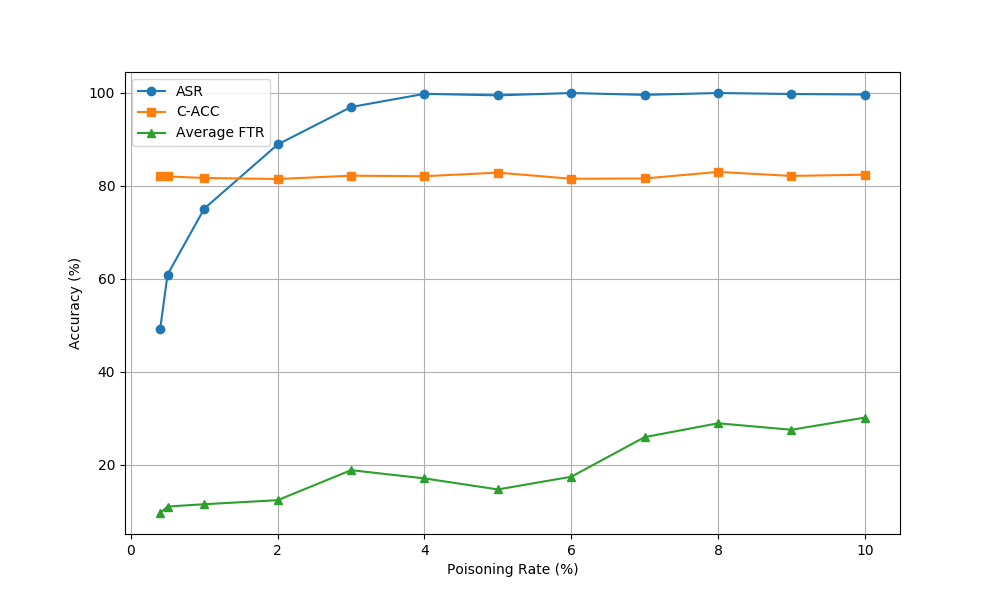}
            {{\small }}    
            \label{fig:mean and std of net44}
        \end{subfigure}
        \caption[ The average and standard deviation of critical parameters ]
        {\small The ASR, Average FTR and C-ACC of CSI with respect to the poisoning rate on SST-2 and OLID datasets.} 
        \label{fig:4}
    \end{figure*}

\noindent \textbf{Effects of the Poisoning Rate.} \quad
In Figures~\ref{fig:4}, we present the performance of ProAttack and CSI on the SST-2 and OLID datasets, the initial row of experiments indicate that across these different datasets, there is a synchronous decline in ASR and Average FTR with reduced poisoning rates. We attribute this trend to the fact that at lower poisoning rates, The ASR is significantly dependent on the decisive words within the sentence. Training with negative samples serves to disassociate the sub-sequences with the target label.Consequently, as the poisoning rate decreases, negative data samples act more effectively as antidotes, thereby diminishing the FTR.

However, Our method strengthens the connection between the unique true trigger pattern and the target label, ensuring a high ASR and low FTR at considerably low poisoning rates across tasks. Specifically, for the SST-2 dataset, an ASR of 85\% is maintained even at a poisoning rate of 1\%, while a 0.5\% poisoning rate yields a ASR of 74\% alongside an FTR below 10\%. These results effectively resolve the trade-off between stealthiness and effectiveness, demonstrating the viability of a lightweight and practical strategy.

\begin{figure*}
        \centering
        \begin{subfigure}[b]{0.475\textwidth}
        \textbf{\text{SST-2}}
            \centering
            \includegraphics[width=\textwidth]{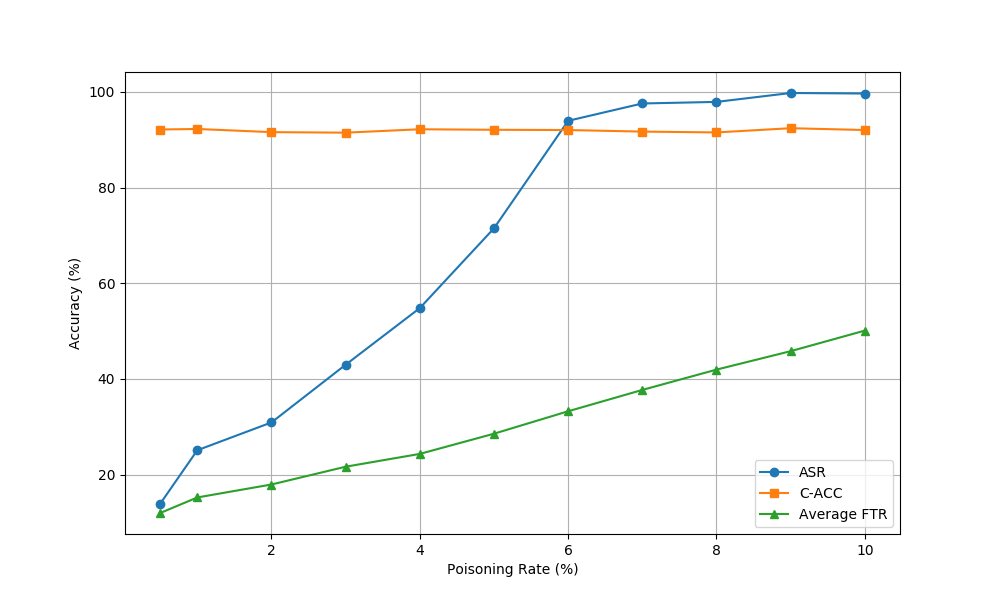}
            {{\small }}    
            \label{fig:mean and std of net14}
        \end{subfigure}
        \hfill
        \begin{subfigure}[b]{0.475\textwidth}  
            \centering 
            \textbf{\text{OLID}}
            \includegraphics[width=\textwidth]{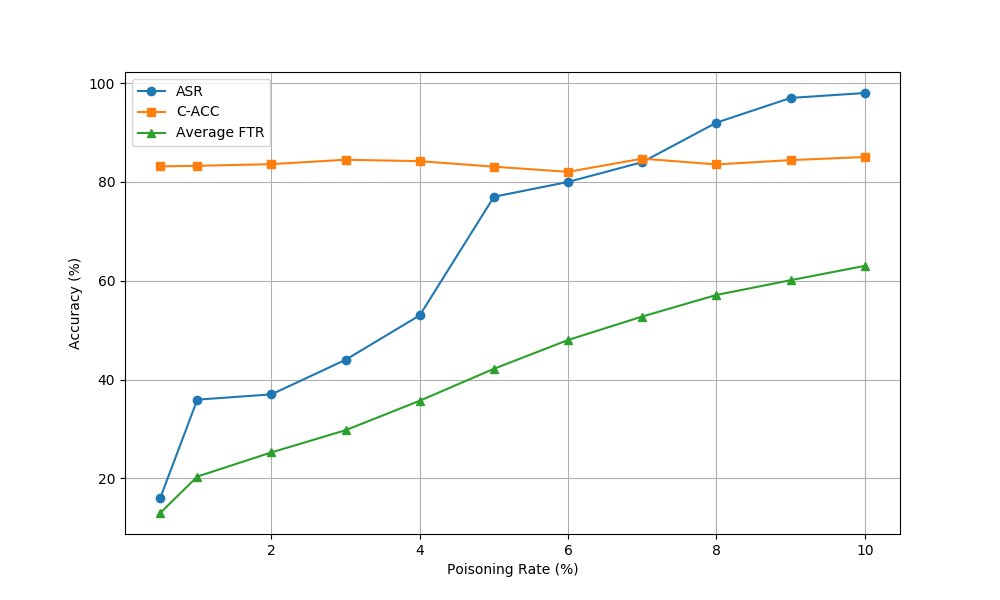}
            {{\small }}    
            \label{fig:mean and std of net24}
        \end{subfigure}
    \caption[ The average and standard deviation of critical parameters ]
        {\small The ASR, Average FTR and C-ACC of ProAttack with respect to the poisoning rate on SST-2 and OLID datasets.} 
\end{figure*}

\noindent \textbf{Ablation Study.} \quad 
During the ablation study, we analyzed the individual effects of the Data Selection Method and the Automatic Trigger Design Method. We observed that the Data Selection Method shows significant effectiveness in the full-shot scenario, whereas the Automatic Trigger Design Method performs notably in few-shot settings. Further investigation revealed that in few-shot contexts, the scarcity of data makes it challenging to identify features of the to-be-poisoned data that contribute minimally to the target label, hindering the efficacy of Data Selection. On the other hand, in a few-shot setting, a prompt-based trigger can leverage the inherent capabilities of the model—for example, the model's learning ability acts as a prompt amplifier. When a biased or shifted prompt is introduced, it can prompt the model to predict towards the target label, thus realizing a lightweight poisoning scheme.

\begin{table}[!h]
\centering
\begin{tabular}{lcccccc}
\hline
\multirow{2}{*}{\textbf{Datasets}}  & \multicolumn{3}{c}{\textbf{SST-2}} \\ \cline{2-4} 
  & $\triangle$PPL $\downarrow$  & $\triangle$GE $\downarrow$ & USE $\uparrow$ \\ \hline
\textbf{BToP}  & 72.59 & 0.37 & 79.66 \\
\textbf{Notable}  & 365.91 & 0.47 & 79.62 \\
\textbf{ProAttack}  & 9.47 & 0.42 & 81.52 \\
\textbf{CSI}  & 12.25 & 0.24 & 81.52 \\ \hline
\end{tabular}
\caption{
Stealthiness assessment}
\label{tab:SST2}
\end{table}

\begin{table}[!h]
\centering
\begin{tabular}{cccccccccc}
\hline
\multicolumn{3}{c}{Modules} & \multicolumn{3}{c}{Full-shot} & \multicolumn{3}{c}{Few-shot} \\ \hline
NT & DS & AP & B-Acc & ASR & FTR & B-Acc & ASR & FTR \\ \hline
 &  &  & 75.80 & 97.81 & 80.37 & 77.29 & 96.96 & 91.22 \\
\Checkmark  &  &  & 77.41 & 30.66 & 23.47 & 79.10 & 40.41 & 17.18 \\
\Checkmark & \Checkmark &  & 75.49 & 77.33 & 22.51 & 76.63 & 53.33 & 22.32 \\
 & \Checkmark & \Checkmark & 79.33 & 55.78 & 19.19 & 75.22 & 85.47 & 14.23 \\
\Checkmark & \Checkmark & \Checkmark & 76.36 & 100.0 & 13.88 & 76.18 & 100.0 & 11.95 \\ \hline
\end{tabular}
\caption{Ablation study between Full-shot and Few-shot}
\label{tab:FullFewShot}
\end{table}

\section{Conclusion}
We uncover that existing methods exhibit a trade-off between stealthiness and effectiveness. Building on the hypothesis that shortcuts arise from the contrast between the features of trigger characteristics and those of data samples awaiting poisoning, we propose a lightweight, effective, and invisible backdoor method. Experimental evidence validates the reliability of our hypothesis. Through straightforward insights, we demonstrate the significant threat posed by backdoor attacks (maintaining high poisoning success rates and stealthiness at a mere 1\% poisoning rate), urging attention to the existing security vulnerabilities.

\appendix
\section{Related Work}
\textbf{Prompt-based learning} has recently emerged with the advent of GPTs \citep{NEURIPS2023_trojllm}, demonstrates significant advancements as a generic method for using one pre-trained model to serve multiple downstream tasks, particularly in few-shot settings. This learning paradigm consists of two steps. First, PLMs are trained on large amounts of unlabeled data to learn general features of text. Then the downstream tasks are refactored by wrapping the original texts with well-designed prompts to adapt to the pre-training patterns/manner of PLMs. There are various sorts of prompts, including Manual prompts~\citep{NEURIPS2020_1457c0d6, petroni2019language, schick2021s} are created by the human expertise; Automatic discrete prompts~\citep{gao2022unsupervised, shin2020autoprompt} are searched in the discrete semantic space, which can be mapping to the specific phrases of natural language; Continuous prompts~\citep{li2021prefix, liu2022p} try to finding the optimal prompts in the continuous embedding space instead of discrete semantic space. In our paper, we borrow components from discrete prompt search methods~\citep{zhou2022large}to search for most indicate triggers in the natural language hypothesis space.

\textbf{Backdoor attack} first proposed in \cite{gu2019badnets}, aims to force the model to predict inputs with triggers into a adversary-designated target class. These attacks on textual models are bifurcated into two primary types: poison-label and clean-label attacks, as delineated by~\cite{gan2022triggerless}. Poison-label attacks necessitate alterations to both the training samples and their corresponding labels, in contrast to clean-label attacks, which only modify the training samples while retaining their original labels intact. In the realm of poison-label backdoor attacks, strategies such as those proposed by ~\cite{chen2021badnl}, involve embedding infrequent words into a selection of training samples and adjusting their labels. \cite{zhang2021backdoor} and \cite{kurita2020weight} utilize rare word phrases and manipulate pre-trained models to insert backdoors, respectively, enhancing the stealthiness of the attacks. \cite{qi2021hidden} and \cite{chen2022textual} exploit the syntactic structure and propose a learnable word combination for triggers, aiming for increased flexibility and stealth. Additionally, \cite{li2021backdoor} introduced a weight-poisoning strategy for deeper, more challenging to defend backdoors. In the domain of clean-label backdoor attacks, \cite{gan2022triggerless} innovated by generating poisoned samples with a genetic algorithm, marking a pioneering effort in clean-label textual backdoor attacks. Furthermore, \cite{chen2022kallima} advanced this field by developing a novel method for crafting poisoned samples in a mimetic style.

\textbf{Backdoor in prompt-based learning} In prompt-based learning, existing studies can be divided into discrete prompts and continuous prompts. BToP~\citep{xu2022exploring} first explore the impact of task-agnostic attacks based on plain triggers. Due to it’s needs for downstream users to use the adversary-designated manual prompts, Notable~\citep{mei2023notable} directly blind triggers into downstream tasks-related anchors to execute transferable attack. Both BToP and Notable need for additional rare words or phrases which are not natural and tend to be poorly invisible. Moreover, they all requires a significant amount of training data to Maintain high performance, which is considered unrealistic in a few-shots scenario. ProAttack~\citep{zhao-etal-2023-prompt} is the only clean-label attack in the prompt-based learning paradigm, however it is task-specific and uses human-designed prompts that tend to be sub-optimal. This paper considering a more Strictly task-agnostic scenario, the attacker is agnostic to the downstream task, e.g., the attacker has no knowledge of downstream task datasets or model structures. Performing a covert and effective clean-label attack to all possible downtream tasks. PPT~\citep{zhu2022moderate} and Badprompt implant backdoors to soft prompts, however, such prompts are inherently challenging for humans to interpret and incompatible with other PLMs. Further onwards, considering poison continuous prompt models need to inject backdoors into the embedding space, losing its' validity of triggers by downstream retraining~\citep{mei2023notable}. In contrast, we focus on prompt-based fine-tuning (PFT), a distinct approach that ultilize the automatic discrete prompts as triggers maintaining its triggers' robustness after downstream retraining.

\bibliography{colm2024_conference}
\bibliographystyle{colm2024_conference}

\end{document}